# Exemplar-based Generative Facial Editing

Jingtao Guo[a], Yi Liu[a,*], Zhenzhen Qian[a], Zuowei Zhou[a]

[a] Beijing Key Lab of Traffic Data Analysis and Mining, School of Computer and Information Technology, Beijing Jiaotong University, Beijing, China

---

[*] Corresponding author. Tel.: +86 135-2082-4503. E-mail address: yiliu@bjtu.edu.cn

**Abstract**

Image synthesis has witnessed substantial progress due to the increasing power of generative model，especially generative adversarial networks. This paper we propose a novel generative approach for exemplar-based facial editing in the form of the region inpainting. Our method first masks the facial editing region to eliminates the pixel constraints of the original image，so as to give full play to the editing potential of the generative model , then exemplar-based facial editing can be achieved by learning the corresponding information from the reference image to complete the masked region. In additional, we impose the attribute labels constraint to model disentangled encodings in order to avoid undesired information being transferred from the exemplar to the original image editing region. Experimental results demonstrate our method can produce diverse and personalized face editing results and provide far more user control flexibility than nearly all existing methods.

# 1. Introduction

This work investigates the exemplar-based face editing task, aims to edit the attributes or key components of a face based on a given exemplar face image, which has broad application prospects in computational photography, image-based rendering and interactive entertainment. Usage of exemplar images allows more precise specification of desired modifications and produce diverse plausible, personalized results. As shown in Fig. 1, the style and structure of editing region is transferred from the exemplar to the source image.

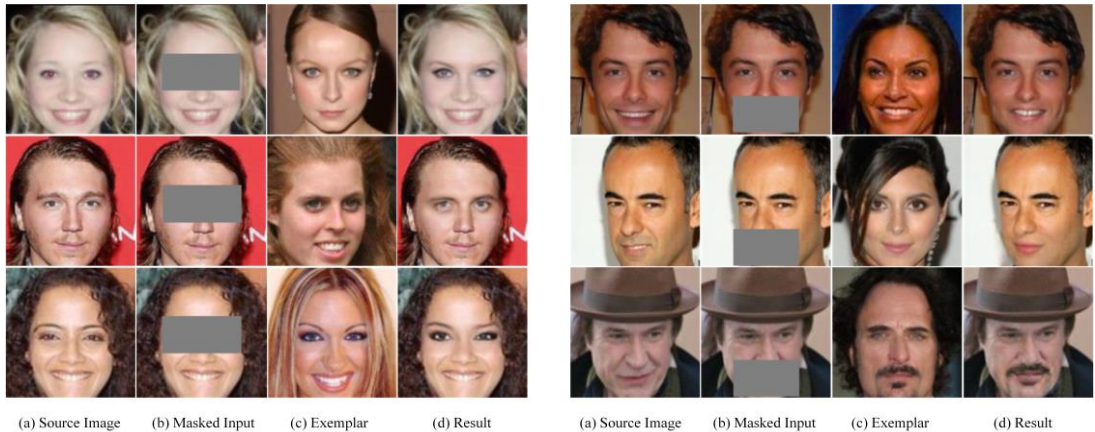

Fig 1. The results of our proposed method for exemplar-based generative facial editing.

Recently, many deep learning based methods [1-7] have emerged and significantly boosted the face editing performance due to the rapid development of deep convolutional networks and generative adversarial networks [8]. These methods formulate face editing as a conditional image generation problem, jointly trained with adversarial networks to synthesis visually realistic and semantically plausible pixels that are consistent with the conditional editing information. StarGAN [1] and AttGAN [2] employ the attribute labels as conditional information to generate a new face with desired attributes while preserving other details. STGAN [3] further improves facial attribute editing quality by introducing selective transfer units to adaptively select and modify encoder feature. However, the information provided by the attribute label is very limited, and cannot express the specific attribute style, for example, the attribute label "eyeglasses" can only indicate whether the face image has eyeglasses, but it cannot provide the specific style information of eyeglasses. So the attribute labels based methods fails to generate complex and diverse, more precise specification of attribute style for facial editing.

To solve this problem, GeneGAN [4] employ both attribute labels and reference images as condition information to perform fine-grained editing by transferring the style of attribute from the exemplar. However, GeneGAN only can transfer a single attribute by exemplar. DNA-GAN [5] and ELEGANT [6] can be viewed as extensions of GeneGAN for transferring multiple attributes simultaneously by adopting the iterative training strategy. However, such iterative training strategy cannot effectively learn disentangled attribute information, resulting in limited ability to transfer multiple attributes of exemplar simultaneously. MulGAN [7] creatively imposes the attribute labels constraint on both the latent feature space and the generated image to force the model learn disentangled encodings for each attribute, which can successfully transfer multiple attribute style simultaneously by an exemplar. However, these example-based methods do not essentially get rid of the limitations of attribute labels, it is still difficult to achieve the transfer of attribute-independent information, like the eye shape and style as shown in Fig. 1, this information is beyond the scope of the attribute labels and cannot be transferred from the reference image.

In order to solve these constraints, this work investigates exemplar-based facial editing from region inpainting perspective and presents a EBGAN model. EBGAN takes a face image with the masked editing region as the input and learn the corresponding information from the reference image to complete the masked region, which can transfer the information of the specified region from the exemplar to the source image and achieve diverse and personalized face editing results. In particular, we propose a conditional version Att-EBGAN that can make a handsome selection for the information of the editing region, and only transfer the part of interest. Att-EBGAN encodes the information of the editing region in a disentangled manner in the latent space by imposing the attribute labels constraint on generated image and latent code, then the selective information transfer can be achieved by filtering certain part of encodings using binary attribute labels, as shown in Fig. 2. The user may not want the "mustache" information in the exemplar to be transferred to the female source image, Att-EBGAN can filter out redundant "mustache" information.

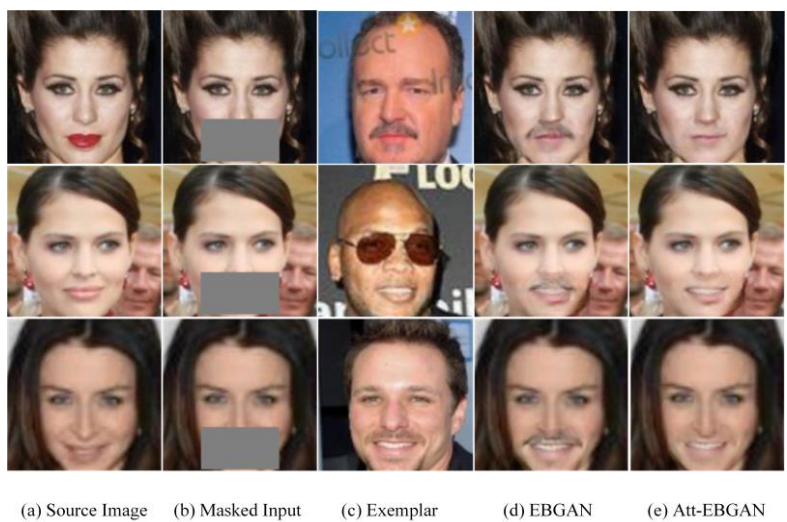

(a) Source Image  (b) Masked Input  (c) Exemplar  (d) EBGAN  (e) Att-EBGAN

Fig 2. The results of EBGAN and Att-EBGAN for exemplar-based generative facial editing, Att-EBGAN can filter out redundant information.

## 3. Method

In this section, we describe the proposed model for exemplar-based facial image editing. Details of the design principles and the loss function for EBGAN and Att-EBGAN are described below.

3.1 *Model*

Let *A* be the ground-truth image with *n* binary attributes $Y^A = \left[ y_1^A, \cdots, y_n^A \right]$. *B* is the exemplar image with the corresponding binary attributes $Y^B = \left[ y_1^B, \cdots, y_n^B \right]$, *B* is generated by randomly shuffling *A*. the binary mask is denoted by *M* (1 for missing regions, 0 for non-missing regions), which is used to mask the facial editing region. Our goal is to enable the generator to learn the corresponding information from the reference image to complete the missing region, so as to achieve example-based facial editing. Figure 2 illustrates an overview of our framework, which consists of four modules described below.

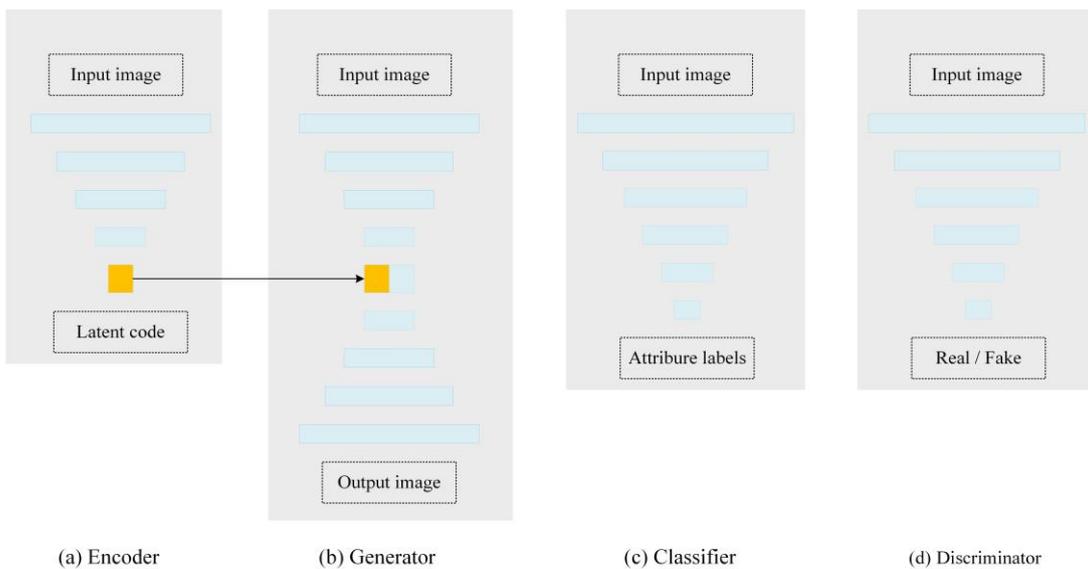

(a) Encoder　　(b) Generator　　(c) Classifier　　(d) Discriminator

Fig. 3. Overview of our method, consisting of four modules. (a) The generator learns to use the latent code of editing region to complete the masked region. (b) The encoder extracts the latent code of editing region. (c) The attribute classifier constrains the completed image to own the correct attributes to assist in modeling disentangled latent code. (d) The discriminator distinguishes between real and fake images.

**Encoder** (Figure 2a). Given a the reference image $B$ and the binary mask $M$, our encoder $E$ generates a latent feature code $Z^B = E(B, M)$. For EBGAN, the latent feature code $Z^B$ are passed directly to the generator $G$ as shown in Fig. 2. For Att-EBGAN, the latent feature code $Z^B$ is constrained by attribute labels before passed to the genenrator. We adopt a similar strategy as MulGAN [7] to encodes the information of editing region in a disentangled manner in the latent space, as shown in Fig. 3. But unlike MulGAN which contain attribute-irrelevant part to extract attribute-independent information. For Att-EBGAN, the latent feature code $Z^B$ is only used to extract the style of attributes in the editing region, and ignoring attribute-irrelevant information. So $Z^B$ is divided into different blocks, where each blocks encodes information of a single attribute: $Z^B = [b_1, \cdots, b_i, \cdots, b_n]$, the $i$-th attribute is encoded into the block $b_i$. Then these blocks are filtered through binary attribute labels $Y^B$: $Z^{B'} = [b_1 \times y_1^B, \cdots, b_i \times y_i^B, \cdots, b_n \times y_n^B]$, where $y_i^A$ is equal to 0, the corresponding attribute-relevant block $a_i$ is set to 0, $y_i^A$ is equal to 1, the corresponding attribute encoding remains unchanged. After this operation, the predefined attribute-related blocks $Z^{B'}$ become a new, stronger representation capability and learnable "labels".

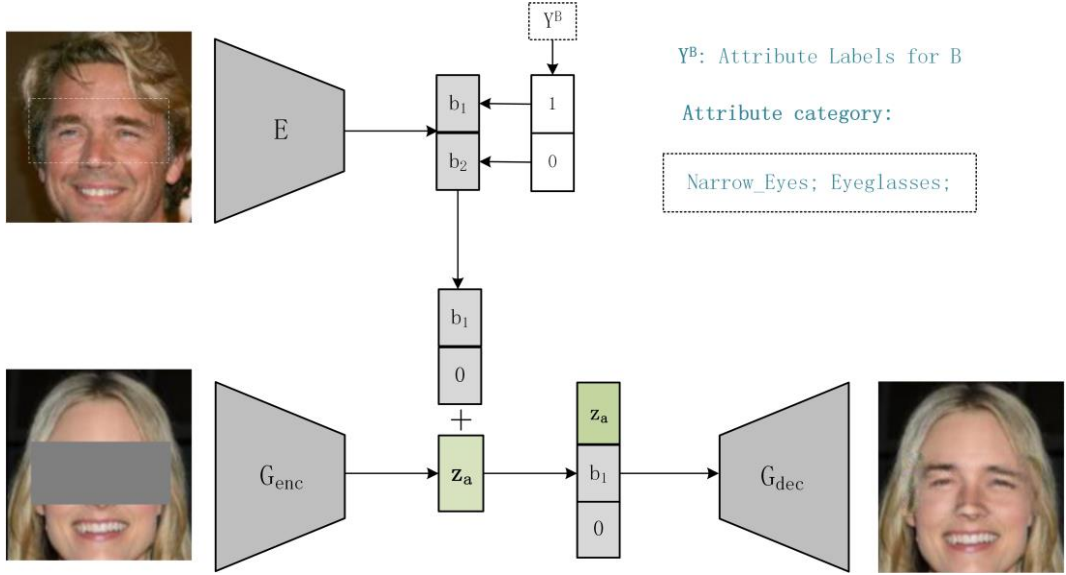

Fig 4. Overview of Att-EBGAN. Attribute label constraints are imposed on the hidden feature space to model disentangled latent code for each attribute.

**Generator** (Figure 2b). Our generator $G$ consists of a encoder ($G_{enc}$) and a decoder ($G_{dec}$). Given a raw image $A$ and the binary mask $M$ for editing region, the input image $\tilde{A}$ is corrupted from the raw image as $\tilde{A} = A \odot (1-M)$, where $\odot$ represents the pixel-wise multiplication. The encoder $G_{enc}$ maps $\tilde{A}$ into latent feature representation $Z^A = G_{enc}(\tilde{A})$. The decoder $G_{dec}$ takes concatenation of $Z^A$ and $Z^{B'}$ as input and outputs the completed image $A^b = G_{dec}(Z^A, Z^{B'})$. Pasting the editing region of $A^b$ to the input image $\tilde{A}$, we get the editing result $y = A^b \odot M + \tilde{A}$. For EBGAN, the attribute disentangled latent code $Z^{B'}$ is replaced with $Z^B$, the others remain unchanged.

**Classifier** (Figure 2c). The classifier $C$ only is used for the Att-EBGAN. As mentioned above, the information of the reference image is encoded in a disentangled way before being sent to generator $G$, when we employ an attribute classifier $C$ to constrain the completed image $A^b$ to own the same attributes with the reference image $B$ in the masked region. which can not only force the separation of attribute-relevant information and attribute-independent information in the masked area of the reference image, but also forces the attribute-relevant information is encoded into the corresponding attribute-relevant blocks for each attribute.

**Discriminator** (Figure 2d). The discriminator $D$ has six down-sampling convolutional layers and a fully-connected layer followed by a sigmoid function to predict whether an image is a real image from the datasets or a fake image produced by $G$, which makes generator $G$ learns how to utilize the reference image information as a semantic guide to produce globally consistent editing results.

3.2 *Loss function*

*Reconstruction Loss*. When the reference image $B = A$, the generator $G$ should learn to best reconstruct the original image $A$ using the attribute disentangled code $Z^{A'}$ for Att-EBGAN. The per-pixel reconstruction loss $L_{rec}$ between the ground-truth image $A$ and the reconstructed one $A^a = G_{dec}(Z^A, Z^{A'})$ is defined as:

$$L_{rec} = \left\| A - A^a \right\|_1 \tag{1}$$

For EBGAN, the attribute disentangled latent code $Z^{A'}$ is replaced with $Z^A$ without attribute label constraints.

*Attribute Classification Loss*. Given input image $\tilde{A}$ and attribute disentangled latent code $Z^{B'}$, we expect the generator $G$ to produce a completed image with the same style as the reference image $B$ in the editing region. To this end, we employ an attribute classifier $C$ and impose the attribute classification loss when optimizing the generator $G$. The attribute classification loss is uesd to optimize the generator $G$ on the generated $A^b$, formulated as follows,

$$L_{cls_g} = \sum_{i=1}^{n} -y_i^B \log C_i(A^b) - (1-y_i^B)\log(1-C_i(A^b)) \qquad (2)$$

where, $C_i(A^b)$ indicates the prediction of the $i$-th attribute for the generated images $A^b$. By minimizing this objective, the generator $G$ can forces the attribute-relevant information is encoded into the corresponding attribute-relevant blocks for each attribute and generate the new image $A^b$ with the same style of attributes in the exemplar $B$.

On the other hand, The attribute classifier $C$ is trained on the source images $A$ with labeled attributes $Y^A$, by the following objective,

$$L_{cls_g} = \sum_{i=1}^{n} -y_i^A \log C_i(A) - (1-y_i^A)\log(1-C_i(A)) \qquad (3)$$

where $C_i(A)$ indicates the prediction of the $i$-th attribute for the ground-truth image $A$. By minimizing this objective, $C$ learns to classify a input image $A$ into its corresponding original attributes $Y^A$.

*Cycle Consistency Loss.* To guarantee that the generated image $A^b = G_{dec}(Z^A, Z^{B'})$ has a similar style as the reference image $B$ in the editing region, we employ the cycle consistency loss and impose reconstruction restrictions on the abstract hidden feature space. First the generated image $A^b$ and the binary mask $M$ are passed to the encoder $E$ to obtain a latent feature code $Z^{\tilde{B}} = E(A^b, M)$ of the generated image $A^b$ in the editing region, then the cycle consistency loss $L_{cyc}$ between the generated image $A^b$ and the reference image $B$ is defined as:

$$L_{cyc} = \left\| Z^B - Z^{\tilde{B}} \right\|_1 \tag{4}$$

*Adversarial Loss.* The adversarial loss is introduced to ensure the editing results visually realistic by adversarial learning between the generator $G$ and the discriminator $D$. Specifically, the adversarial losses for the discriminator ($L_{adv_d}$) and for the generator ($L_{adv_g}$) are formulated as below:

$$L_{adv_d} = -E\left[\log\left(D(x^a)\right) + \log\left(1 - D(y)\right)\right] \tag{5}$$

$$L_{adv_g} = E\left[\log\left(D(y)\right)\right] \tag{6}$$

where $y = A^b \odot M + \tilde{A}$, which replaces the pixels in the editing region of original image with the generated corresponding ones, the discriminator network $D$ tries to distinguish between a real image $A$ and the generated $y$.

To make things clear, the final objective functions to optimize the $G$, $D$ and $C$ networks of Att-EBGAN are listed below:

$$L_G = L_{adv_g} + \lambda_{rec} L_{rec} + \lambda_{cyc} L_{cyc} + \lambda_g L_{cls_g} \tag{7}$$

$$L_D = L_{adv_d} \tag{8}$$

$$L_C = L_{cls_c} \tag{9}$$

Here, $\lambda_{cyc}$, $\lambda_{rec}$ and $\lambda_g$ are weights to define the importance of different losses for the generator network.

For the EBGAN, the final objective functions to optimize the $G$, $D$ are listed below:

$$L_G = L_{adv_g} + \lambda_{rec}L_{rec} + \lambda_{cyc}L_{cyc} \tag{10}$$

$$L_D = L_{adv_d} \tag{11}$$

Here, the classifier $C$ and corresponding classification loss are discarded, The effect of $\lambda_{cyc}$ and $\lambda_{rec}$ are the same as above.

## 4. Experiments

In this section, we first demonstrate the ability of EBGAN to produce a particular editing result with the reference image. Then, we show that Att-EBGAN can achieve the selective information transfer by filtering certain part of encodings using binary attribute labels.

4.1 *Experiment setup*

We evaluate the proposed EBGAN and Att-EBGAN on the CelebA dataset [9], which is consist of 200k facial images, each with 40 attributes annotation. We follow the standard protocol: splitting the dataset into 160k images for training, 20k for validation and 20k for testing, the training and validation set together are used to train our model while the testing set is used for evaluation. The model is trained by Adam optimizer [10] with a batch size of 32 and the learning rate of 0.0001. The coefficients for the losses in Eq. (7) are set as: $\lambda_{cyc}=10$, $\lambda_{rec}=100$ and $\lambda_{g}=10$, which balances the effects of different losses.

*4.2 Results*

*4.2.1 EBGAN*

First, we evaluate the performance of EBGAN in terms of local facial components, which is related to one local key facial part (mouth). The results in Fig. 5 shows the facial image editing results when explicitly controlling the mouth shape, the style and structure of editing region is transferred from the exemplar to the source image.

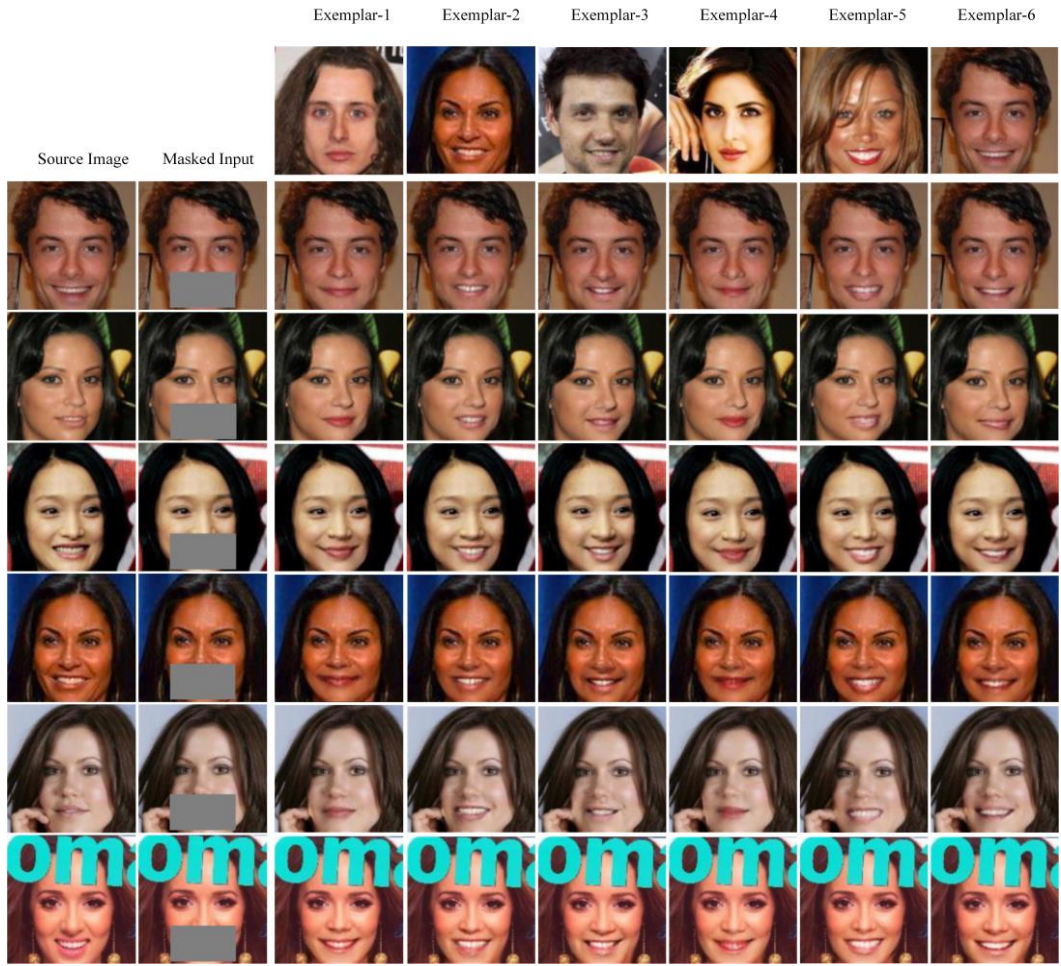

Fig 5. The results of EBGAN for exemplar-based local facial component (mouth) editing

Besides editing single local facial component, our method can also achieve good editing results for multiple facial components simultaneously. Fig. 6 shows the multiple facial components (eyes, eyebrows, nose and mouth) of exemplar are successfully transferred to the source image.

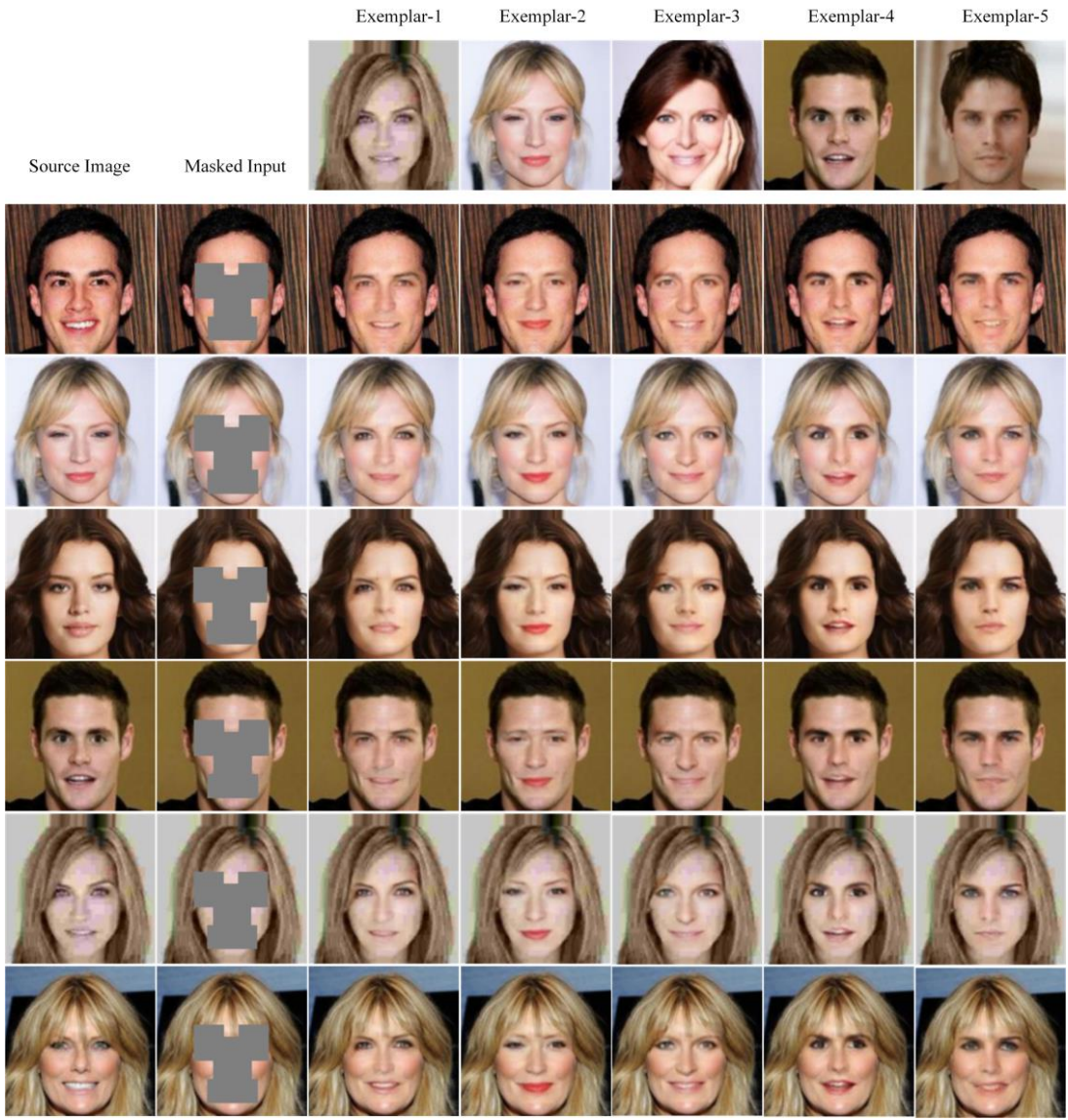

Fig 6. The results of EBGAN for exemplar-based multiple facial components editing simultaneously.

Finally, to demonstrate the powerful example-based editing capabilities of our method, we conducted more challenging experiments that edits entire face. As shown in

Figure 7, EBGAN successfully achieves "face swapping" between the source image and the reference image,

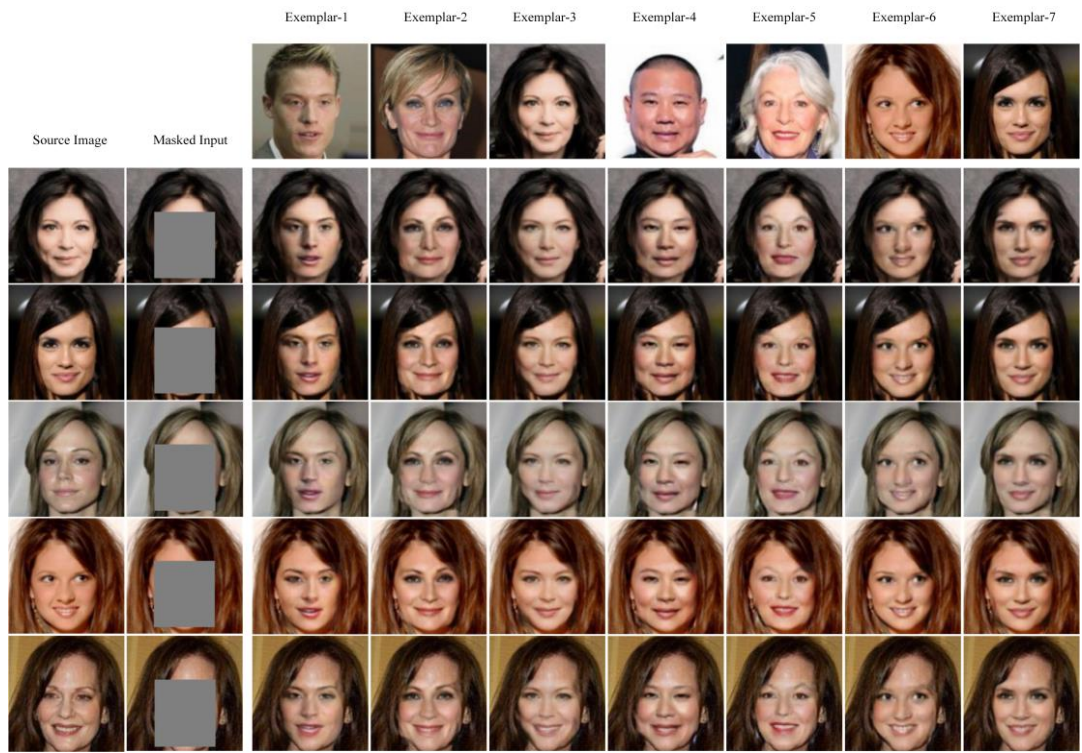

Fig 7. The results of EBGAN for entire facial editing.

*4.2.2 Att-EBGAN*

The Conditional version Att-EBGAN encodes the information of the editing region in a disentangled manner in the latent space by imposing the attribute labels constraint on generated image and latent code. It allows users to control the process of information transfer through binary attribute labels. As shown in Figure 8, Att-EBGAN filters out redundant information "mustache" in the face editing process.

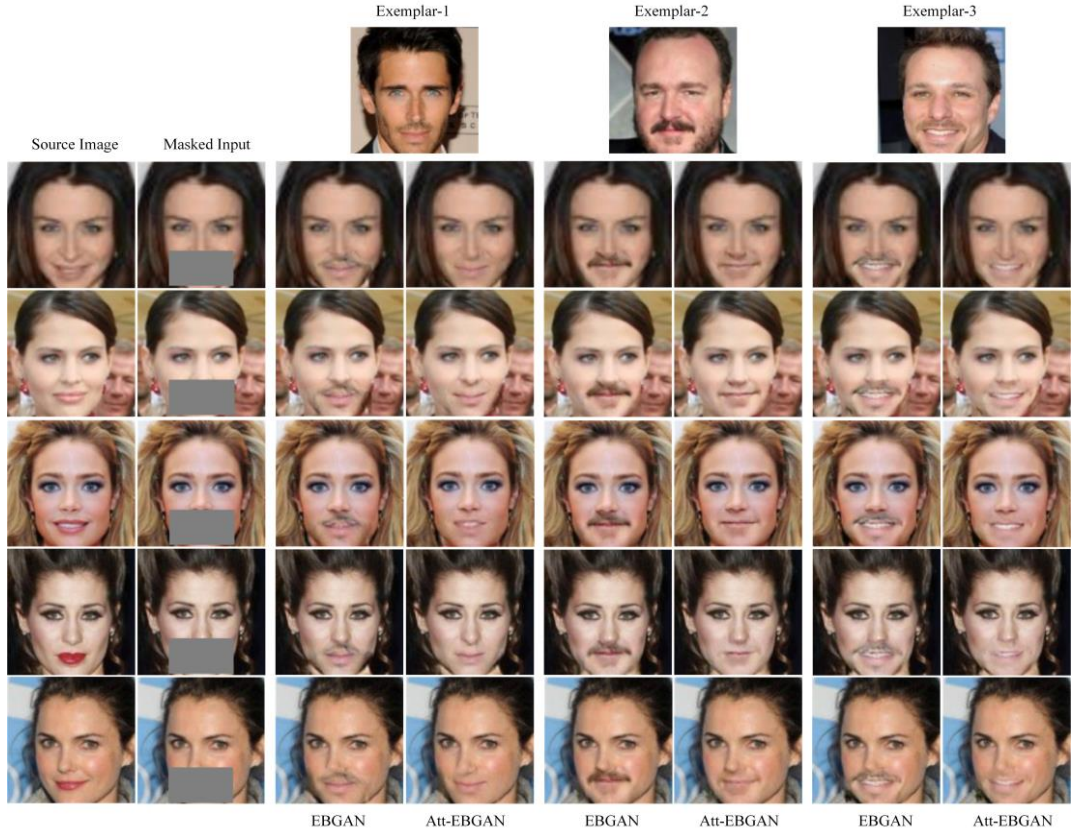

Fig 8. Att-EBGAN filters out redundant information "mustache" in the editing region.

## 5. Conclusion

From the perspective of image inpainting, we propose a novel generative approach EBGAN for exemplar-based facial editing, which can make full use of the information in the reference image to achieve produce diverse and personalized face editing results. Furthermore, we properly consider the relation between the attributes and the latent representation and propose a conditional version Att-EBGAN which incorporates the attribute label constraint to achieve selective information transfer from the exemplar to source image. Experiments demonstrate that our method owns superior performance and provide more flexibility than nearly all existing methods.

## Acknowledgments

The authors acknowledge support from the Natural Science Foundation of China (No.61300072, 31771475).